# Effective Learning of a GMRF Mixture Model

SHAHAF E. FINDER<sup>ⓘ</sup>, ERAN TREISTER<sup>ⓘ</sup>, AND OREN FREIFELD<sup>ⓘ</sup>
Department of Computer Science, Ben-Gurion University of the Negev, Be'er Sheva 84105, Israel

Corresponding author: Shahaf E. Finder (finders@post.bgu.ac.il)

This work was supported in part by the Israeli Council for Higher Education (CHE) via the Data Science Research Center, Ben-Gurion University of the Negev, Israel; and in part by the Lynn and William Frankel Center for Computer Science at BGU. The work of Shahaf E. Finder was supported by the Kreitman School of Advanced Graduate Studies. The work of Oren Freifeld was supported in part by the Israel Science Foundation Personal under Grant 360/21.

**ABSTRACT** Learning a Gaussian Mixture Model (GMM) is hard when the number of parameters is too large given the amount of available data. As a remedy, we propose restricting the GMM to a Gaussian Markov Random Field Mixture Model (GMRF-MM), as well as a new method for estimating the latter's sparse precision (*i.e.*, inverse covariance) matrices. When the sparsity pattern of each matrix is known, we propose an efficient optimization method for the Maximum Likelihood Estimate (MLE) of that matrix. When it is unknown, we utilize the popular Graphical Least Absolute Shrinkage and Selection Operator (GLASSO) to estimate that pattern. However, we show that even for a single Gaussian, when GLASSO is tuned to successfully estimate the sparsity pattern, it does so at the price of a substantial bias of the values of the nonzero entries of the matrix, and we show that this problem only worsens in a mixture setting. To overcome this, we discard the nonzero values estimated by GLASSO, keep only its pattern estimate and use it within the proposed MLE method. This yields an effective two-step procedure that removes the bias. We show that our ''debiasing'' approach outperforms GLASSO in both the single-GMRF and the GMRF-MM cases. We also show that when learning priors for image patches, our method outperforms GLASSO even if we merely use an educated guess about the sparsity pattern, and that our GMRF-MM outperforms the baseline GMM on real and synthetic high-dimensional datasets.

**INDEX TERMS** Gaussian mixture model, GMRF, sparse inverse covariance matrix, probabilistic models.

## I. INTRODUCTION

The ubiquitous Gaussian Mixture Model (GMM) (*e.g.*, see [1]) is widely used for modeling complex non-unimodal distributions, and has found applications in numerous fields including, but not limited to, computer vision and image processing [2]–[9], signal processing [10]–[13], speech recognition [14], [15], anomaly detection [16], [17] and biology [18], [19]. The number of parameters in each Gaussian, however, is quadratic in $n$, the dimension of the data, and thus, when there is not enough data to support an effective estimation, the GMM is prone to become over-parameterized [20], [21], hindering the applicability of the model. Moreover, in certain settings (though not always), $K$, the number of Gaussians in the mixture, also tends to grow with $n$, making the over-parameterization problem worse since the average number of points per Gaussian decreases when $K$ increases. For example, Zoran and Weiss [6], who learned a GMM over 8-by-8 image patches (*i.e.*, $n = 64$), reported that $K$ needed to be as high as 200 to produce good results. Other scenarios (*e.g.*, larger image patches) might require even a higher $K$. That said, there also exist of course high-dimensional cases where a low $K$ suffices.

To overcome the over-parameterization problem, we propose reducing the number of parameters by restricting each of the components of the GMM to be a Gaussian Markov Random Field (GMRF), thereby obtaining a GMRF Mixture Model (GMRF-MM). A GMRF is a Gaussian whose precision matrix (namely, inverse covariance), denoted by $Q$, is sparse. Such a $Q$ is associated with a probabilistic graphical model [22], [23]; we will elaborate on this well-known fact in § II.

*Remark 1:* Unless a block-diagonal structure is involved, a sparse covariance matrix usually does not imply a sparse precision matrix and vice verse. In particular, in the context of Gaussians, while zeros in the covariance matrix imply independence between the relevant variables, zeros in the precision matrix do *not* imply such independence; rather, they imply **conditional** independence (see § II).

The associate editor coordinating the review of this manuscript and approving it for publication was Massimo Cafaro<sup>ⓘ</sup>.

  

*Remark 2:* The GMRF-MM should not be confused with models that combine a (typically low-dimensional) GMM with a discrete-state Markov Random Field (MRF) over the spatially- (or temporally-) dependent labels associated with the observations (as was done in, e.g., [24], [25]).

The GMRF assumption, common in the single-Gaussian case (see [23], [26] and references therein), is rarely used in the case of mixtures. This raises a natural question: how come the idea of reducing the number of parameters by restricting the GMM to a GMRF-MM has been hardly explored so far? We suspect that a plausible explanation for this is that traditional methods for estimating the sparse $Q$ (in the single-Gaussian case, which is an indispensable building block in the mixture setting) are not effective enough, especially when the sparsity pattern (*i.e.*, the locations of the zero entries of $Q$) is unknown. For example, consider the widely-used Graphical Least Absolute Shrinkage and Selection Operator (GLASSO) [27], which is based on an $\ell_1$ penalty (see also [28]–[32], and other related solvers based on $\ell_1$ that approximate GLASSO [33]–[35]). In this paper we show that even when GLASSO's regularization parameter is tuned to successfully estimate the sparsity pattern, GLASSO substantially biases the estimated values of the nonzero entries of $Q$. In fact, we show this undesirable effect is present regardless whether one excludes the diagonal terms in $Q$ from the $\ell_1$ penalty [36] or not. Moreover, and especially important in the context of our paper which focuses on the mixture setting, we show that *the negative effect of that bias on the resulting model in the GMM case is even worse than in the single-Gaussian case*. Thus, the seemingly-natural choice of GMM+GLASSO [37]–[41] might lead to poor estimates, rendering the GMRF-MM impractical. In fact, we show that GMM+GLASSO can obtain worse results than the baseline (*i.e.*, non-regularized) GMM even when the number of data points is low.

Note that whenever possible, and because of the aforementioned bias, it is beneficial to learn a GMRF-MM using a pre-defined sparsity. Indeed, in many applications the sparsity pattern is either known or can be assumed via an educated guess; *e.g.*, this is the case in many successful computer-vision Markov Random Field models [42], [43] or in Geoscience where regions in geographical maps are often assumed to exhibit a Markovian structure [44]. In this work we propose a framework for an effective and efficient learning a GMRF-MM. In the case when the sparsity can be pre-defined, we propose to estimate the sparse $Q$ using a Newton-type optimization method for finding the appropriately-constrained Maximum Likelihood Estimate (MLE). When the sparsity pattern itself must be estimated as well, we employ a two-step procedure. First, we solve the GLASSO problem; this can be done using a variety of available methods mentioned earlier. Next, we *discard the GLASSO estimates* of the nonzero entries of $Q$ because these are usually *substantially biased* and keep only the GLASSO estimate of the sparsity pattern. To find the "debiased" MLE, we use that Newton-type method given the estimated sparsity pattern. This "debiasing" procedure effectively removes the GLASSO bias.

Our experiments show that the proposed method outperforms GLASSO in both the single-Gaussian and the GMRF-MM cases. We also show that when learning a GMM over image patches, the proposed method outperforms GLASSO in both of the mutually-exclusive cases: 1) when we merely make an educated guess about the sparsity pattern; 2) when we use sparsity pattern estimated by GLASS but debias the GLASSO estimates of the nonzero values. In addition, we show that, with the proposed method, a GMRF-MM outperforms a baseline GMM on real and synthetic high-dimensional datasets.

To summarize, **our key contributions are as follows.**
noitemsep

1) We propose a framework for learning a GMM that overcomes the over-parameterization problem by assuming each component is a GMRF.
2) When the sparsity pattern, $\Omega$, is known, we propose an efficient optimization method for finding the MLE of $Q$. When $\Omega$ is unknown, we use 2-step procedure that *debiases* (*i.e.*, removes the bias of) the GLASSO estimate.
3) We show that estimating the GMRF-MM with the debiased GLASSO lets us learn an effective GMRF-MM of unknown sparsity patterns ($\Omega$ is component-specific; *i.e.*, each Gaussian in the mixture may have a different sparsity pattern).
4) We provide new theoretical results that further explain the GLASSO bias (with either a penalized or non-penalized diagonal).

Our code is publicly available at https://github.com/shahaffind/GMRF-MM.

## II. PRELIMINARIES

A $K$-component GMM in $\mathbb{R}^n$ is given by the following probability density function (pdf):

$$p(x; \Theta) = \sum_{k=1}^{K} \pi_k \mathcal{N}(x; \mu_k, \Sigma_k) \quad x \in \mathbb{R}^n \quad (1)$$

where $\Theta = \{\mu_k, \Sigma_k, \pi_k\}_{k=1}^{K}$, $\mu_k \in \mathbb{R}^n$ and $\Sigma_k \in \mathbb{R}^{n \times n}$ are the mean and (symmetric positive-definite) covariance matrix of Gaussian $k$, the weights $(\pi_k)_{k=1}^{K}$ form a convex combination, and

$$\mathcal{N}(x; \mu, \Sigma) \triangleq \frac{1}{(2\pi)^{n/2}|\Sigma|^{1/2}} \exp\left(-\tfrac{1}{2}\|x - \mu\|_{\Sigma^{-1}}^2\right) \quad (2)$$

where

$$\|x - \mu\|_{\Sigma^{-1}}^2 \triangleq (x - \mu)^T \Sigma^{-1} (x - \mu). \quad (3)$$

Given $N$ *i.i.d.* observations, $(x_i)_{i=1}^{N} \subset \mathbb{R}^n$, drawn from a GMM parameterized by $\Theta$, the MLE of $\Theta$ is

$$\widehat{\Theta}_{\text{MLE}} = \arg\max_{\Theta} \prod_{i=1}^{N} \sum_{k=1}^{K} \pi_k \mathcal{N}(x; \mu_k, \Sigma_k). \quad (4)$$

A closed-form solution for Eq. (4) does not exist. Thus, a popular solution is to introduce auxiliary variables

(called hard assignments or labels) and to use *an* Expectation-Maximization (EM) algorithm, which works iteratively by alternating between two steps:

1) An **E step** where, for each $i \in \{1, \ldots, N\}$ and each $k \in \{1, \ldots, K\}$, $w_{ki}$, the soft assignment of data point $x_i$ to Gaussian $k$, is computed;
2) An **M step**, which maximizes, w.r.t. $\Theta$, the conditional expectation of the so-called complete-data log-likelihood given those soft assignments.

For more details see [21]. Let $Q_k = \Sigma_k^{-1}$ denote the precision matrix of Gaussian $k$. It can be shown that, during the M step, the estimation of $\Sigma_k$, the covariance of Gaussian $k$, is equivalent to the MLE of the covariance in a single Gaussian while also taking the soft assignments into account by using them as weights. Due to the invariance property of the MLE [45], we can also reformulate this in terms of the (weighted) MLE of $Q_k$. That is, given the (weighted) empirical covariance matrix

$$S_k = \frac{1}{\sum_i w_{ki}} \sum_{i=1}^{N} w_{ki}(x_i - \hat{\mu}_k)(x_i - \hat{\mu}_k)^T \quad (5)$$

(where $\hat{\mu}_k = \frac{1}{\sum_i w_{ki}} \sum_{i=1}^{N} w_{ki} x_i$ is the weighted sample mean of Gaussian $k$), the MLE of $Q$ is obtained by minimizing the negative log-likelihood (for a single Gaussian, and when discarding constants),

$$\mathcal{L}(Q_k) = -\log \det Q_k + \text{tr}(Q_k S_k) \quad (6)$$

(where tr() denotes a matrix trace) w.r.t. $Q_k$. Importantly, the optimization is done subject to the constraint that $Q_k \succ 0$ (*i.e.*, $Q_k$ is symmetric positive-definite). Since the function in Eq. (6) is convex, a global minimum is found by setting the gradient

$$\nabla \mathcal{L}(Q_k) = S_k - Q_k^{-1} \quad (7)$$

to zero, leading to the known result,

$$\hat{Q}_{\text{MLE}} = S_k^{-1}. \quad (8)$$

This closed-form estimator, however, is oblivious to possible constraints on, or a regularization over, $Q_k$. Particularly, a regularization is necessary when $S_k$ is rank deficient, in which case the closed-form estimator in Eq. (8) is undefined (since the matrix is not invertible). Such a situation can occur during the EM process, especially when the number of data samples is small.

### A. GMRF AND GRAPHICAL LASSO

Let $x = (x_1, \ldots, x_n) \sim \mathcal{N}(\mu, \Sigma)$. Define a graph $\mathcal{G} = (V, E)$, whose nodes are $V = \{1, \ldots, n\}$ and its (undirected) edges, $E$, are defined such that $(i, j) \notin E$ if and only if $x_i \perp x_j | x_{-ij}$ where $x_{-ij} = \{x_k | k \notin \{i, j\}\}$. The notation $x_i \perp x_j | x_{-ij}$ indicates that conditioned on the variables in all the *other* nodes, the two variables, $x_i$ and $x_j$, are conditionally independent. We then say that $x$ is a GMRF w.r.t. $\mathcal{G}$. It can be shown [23] that the precision matrix, $Q = \Sigma^{-1}$, satisfies

$$x_i \perp x_j | x_{-ij} \iff Q_{ij} = 0. \quad (9)$$

*Remark 3:* Note that Eq. (9) is a different statement from

$$x_i \perp x_j \iff \Sigma_{ij} = 0, \quad (10)$$

which says that two entries of a Gaussian random vector are independent if and only if they are uncorrelated (namely, have zero covariance).

Thus, assuming that $Q$ is sparse implies a probabilistic graphical model with a statistical interpretation that can be justified, either exactly or approximately, in many applications. The sparsity of the graph, however, adds a constraint that must be addressed during the process of the estimation of $Q$. Particularly, the closed-form solution in Eq. (8) is no longer applicable as it usually violates that constraint. When the sparsity pattern $\Omega$ is known, we can use gradient-based optimization methods that constrain the minimizer to $\Omega$; *i.e.*, such a method would solve

$$\arg\min_{Q \succ 0} \mathcal{L}(Q) \text{ subject to } \text{Supp}(Q) \subseteq \Omega \quad (11)$$

where $\text{Supp}(Q) = \{(i,j) \mid Q_{ij} = Q_{ji} \neq 0\}$, and $\Omega \subsetneq (1, \ldots, n) \times (1, \ldots, n)$ is a known set of possible nonzero entries. We will return to this problem in § III where we propose an efficient method to solve it. When $\Omega$ is unknown, a popular approach is to use the GLASSO estimator [27]:

$$\arg\min_{Q \succ 0} \mathcal{L}(Q) + \lambda \|Q\|_1 \quad (12)$$

where $\|\bullet\|_1$ is the element-wise matrix $\ell_1$ norm, and $\lambda > 0$ is a regularization parameter. The addition of the $\ell_1$ penalty promotes the sparsity of the minimizer, thus implying a graphical model according to Eq. (9), freeing the user from having to know or assume the sparsity pattern. A common variant of Eq. (12) does not include the diagonal terms on $Q$ in the $\ell_1$ penalty term [36], to reduce the bias of the diagonal terms which are known to be non-zero. As we will show, however, this method has a main drawback.

*Remark 4:* While the choice of $K$ itself is beyond the scope of our paper, we note that the proposed method can be used as is within existing model-selection paradigms. That is, in the first stage one fits multiple models to the data, each of them using a different value of $K$ over some predefined range. Next, an unsupervised metric such as the Silhouette score [46] is used to determine a good trade-off between the goodness of fit and penalizing high values of $K$.

### III. METHOD

We now describe the estimation process of a GMRF-MM. Generally, we use the EM algorithm described in § II, but add constraints or regularization terms to yield sparse $Q_k$ in the MLE estimation (the M-step). To simplify the notation, in the remainder of this section we drop the subscripted $k$ (*e.g.*, we will write $Q$, $\Sigma$ and $S$ instead of $Q_k$, $\Sigma_k$ and $S_k$).

When adding a constraint or a penalty over $Q$, usually there is no longer a closed-form solution for the minimizer of Eq. (6) during the M-step. Particularly, this happens in the two cases considered here: the minimization problem in Eq. (11), when the sparsity pattern of $Q$ is assumed to be known, and in Eq. (12) (*i.e.*, GLASSO), when that pattern is unknown and an $\ell_1$ regularization is used to promote sparsity. In both these cases, we handle the minimization via an appropriate gradient-based iterative method. Any such method first computes a search direction that involves the gradient from Eq. (7), and then takes a step in that direction. That is, the update at iteration $t$ is given by

$$Q^{(t)} = Q^{(t-1)} + \alpha^{(t)} \Delta^{(t)}, \qquad (13)$$

where $\Delta^{(t)}$ is the search direction and $\alpha^{(t)}$ is the step size, usually obtained by a line search. For example, if we use Gradient Descent to solve Eq. (6), we have $\Delta^{(t)} = -\nabla \mathcal{L}(Q^{(t-1)})$. In order to impose a given sparsity pattern $\Omega$ on the solution, it is possible to project the gradient-descent direction to $\Omega$. To solve the GLASSO problem (Eq. (12)) one may use a solver that is based on the proximity operator (known as "soft-thresholding") to handle the $\ell_1$ penalty. Below we describe these approaches in more detail.

### A. ESTIMATING $Q$ WITH A KNOWN SPARSITY PATTERN

Suppose that we wish to solve Eq. (11). One main problem arises in this case: any gradient-based approach requires the computation of $Q^{-1}$ at each step. This is expensive since $Q^{-1}$ is usually a dense matrix even if $Q$ itself is sparse. Moreover, if we wish to project the gradient to a given sparsity pattern (compute arbitrary entries from the gradient), there is no easy way to compute only the relevant entries of $Q^{-1}$ without computing all the entries first. An exception of this is when the underlying graph of $Q$ is assumed to be chordal; see the following remark for details.

*Remark 5:* When the underlying graph of $Q$ is assumed to be chordal, it is possible to compute the entries $(i, j)$ of $Q^{-1}$ where $Q_{ij} \neq 0$ efficiently [47]. Similarly, if the pattern in $\Omega$ is chordal, it is possible to solve Eq. (11) by a recursive elimination. In addition, one can extend the allowed nonzero pattern of $Q$ to be chordal [48], and use the extended pattern to approximate the solution of Eq. (11) for non-chordal graphs. Either way, that is not the approach we follow here, since the general graphs that we consider are non-chordal, and while their associated matrices are fairly large, they are still small enough to be inverted at a reasonable cost.

When the optimization problem in Eq. (6) is ill-conditioned, first-order methods might require numerous iterations, and each such iteration requires the inversion operation. To avoid the repeated matrix inversions in such problems, it is common to solve the optimization problem by using a second-order approximation for obtaining the Newton descent direction, using the $n^2$-by-$n^2$ Hessian

$$\nabla^2 \mathcal{L}(Q) = Q^{-1} \otimes Q^{-1} \qquad (14)$$

where $\otimes$ is the Kronecker product. By definition, the computation of $Q^{-1}$ from the gradient in Eq. (7) can be reused for the Hessian in Eq. (14) – this fact is a main advantage of second-order methods. The Newton direction $\Delta$ for Eq. (11) is obtained by solving the *projected Newton problem*:

$$\min_{\Delta: \text{Supp}(\Delta) \subseteq \Omega} \text{tr}(\nabla \mathcal{L}(Q) \Delta) + \tfrac{1}{2} \text{tr}(\Delta Q^{-1} \Delta Q^{-1}). \qquad (15)$$

The Hessian matrix (Eq. (14)) might be dense and large, but multiplying it with a sparse descent direction $\Delta$ and projecting the result to $\Omega$ can be obtained efficiently [29], [49]. Particularly, the Hessian matrix never needs to be explicitly formed or stored in memory. This allows us to efficiently solve Eq. (15) for a given sparsity pattern $\Omega$ using an iterative method. The direction can be found by setting the gradient of Eq. (15) to 0, leading to the following equation:

$$Q^{-1} \Delta Q^{-1} = -\nabla \mathcal{L}(Q). \qquad (16)$$

To solve it we use the projected and Preconditioned Conjugate Gradient (PCG) method [50], which is relatively insensitive to a high condition number; note that the conditioning of the problem (the eigenvalues of the Hessian Eq. (14)) depends on the estimated matrix $Q$ itself, and can be arbitrarily high.

Algorithm 1 presents the method for estimating the MLE subject to a given support. The projected Newton problem (Eq. (15)) is solved in the inner loop of the algorithm, where the projected and Preconditioned Conjugate Gradient (`projPCG`) method is applied iteratively, to solve the linear Newton problem (Eq. (16)) subject to the given support. In this method, each PCG step (which mostly includes the multiplication of the Hessian with the previous direction) is projected to the given support to maintain the desired sparsity pattern. After calculating the descent direction $\Delta$, we perform an Armijo line search to find the optimal step size. The line search also verifies the descent result is an SPD (symmetric positive definite) matrix.

The preconditioner $M$ that we use inside `projPCG` is diagonal; *i.e.*, it has a weight for each entry of $\Delta_{ij}$. We define the weight $M_{ij}$ for the entry $(i, j)$

$$M_{ij} = \begin{cases} Q_{ii}^{-1} Q_{jj}^{-1} + (Q_{ij}^{-1})^2, & i \neq j \\ (Q_{ii}^{-1})^2, & i = j. \end{cases} \qquad (17)$$

---

**Algorithm 1** Estimating $Q$ With a Known Support

**Input:** $S$, $\Omega$, $Q^{(0)}$
    **for** $t = 1, 2, \ldots$ **do**
        $\widehat{G}^{(t)} \leftarrow S - (Q^{(t-1)})^{-1}$
        $G^{(t)} \leftarrow$ Project $\widehat{G}^{(t)}$ using the support $\Omega$
        $\Delta^{(t)} \leftarrow$ projPCG$(Q^{(t-1)}, G^{(t)}, \Omega)$
        $\alpha^{(t)} \leftarrow$ Armijo line-search$(Q^{(t-1)}, G^{(t)}, \Delta^{(t)})$
        $Q^{(t)} \leftarrow Q^{(t-1)} + \alpha^{(t)} \Delta^{(t)}$
    **end for**
    **return** $Q^{(t)}$

---

### B. ESTIMATING $Q$ WITH AN UNKNOWN SPARSITY PATTERN

If $\Omega$ is unknown a-priori, then one of the popular options for finding it is the GLASSO (Eq. (12)). To solve this problem,

the proximal Newton method is often used together with an active set approach [29], [49]. That is, we first restrict the Newton update at the $k$-th iteration to the support of the "free set":

$$\mathcal{F}(\boldsymbol{Q}) = \{(i, j) : \boldsymbol{Q}_{i,j} \neq 0 \lor |\boldsymbol{S}_{i,j} - (\boldsymbol{Q}^{-1})_{i,j}| > \lambda\}. \quad (18)$$

Given the free set $\mathcal{F}(\boldsymbol{Q})$, we apply the quadratic approximation to the smooth part of the objective $\mathcal{L}(\boldsymbol{Q})$, while keeping the $\ell_1$ term intact. This results in a LASSO problem instead of Eq. (15):

$$\min_{\boldsymbol{\Delta}:\mathrm{Supp}(\boldsymbol{\Delta})\subseteq\mathcal{F}} \mathrm{tr}(\nabla\mathcal{L}(\boldsymbol{Q})\boldsymbol{\Delta}) + \tfrac{1}{2}\mathrm{tr}(\boldsymbol{\Delta}\boldsymbol{Q}^{-1}\boldsymbol{\Delta}\boldsymbol{Q}^{-1}) \\ + \lambda \|\boldsymbol{Q} + \boldsymbol{\Delta}\|_1. \quad (19)$$

This LASSO problem can be solved using Coordinate Descent [51] or the (projected and preconditioned) Conjugate Gradient for the LASSO problem [52], [53]. Once the LASSO problem is solved, a line search is applied. In the EM framework, we can use any GLASSO solver available from the literature. Specifically we use a Newton-PCG algorithm that follows similar lines as the algorithm in the previous section, since it is the best algorithm known to us for a parallel GPU implementation at the scales that we consider.

### C. DEBIASING THE GLASSO

As stated earlier, the GLASSO uses an $\ell_1$ regularization to promote sparsity. Although it achieves that goal, it also *penalizes the magnitude of nonzero entries* of $\boldsymbol{Q}$. We refer to this problem as the GLASSO bias. In fact, as we demonstrate later (see § IV), this bias is even more damaging in the mixture setting than in the single-Gaussian case. This is because of the following vicious cycle: the bias in the estimates of the $\boldsymbol{Q}_k$'s yields poor estimates of the $w_{ki}$ weights during the E-step of the EM algorithm, and these, in turn, yield poorer estimates of the $\boldsymbol{Q}_k$'s in the M-step, and so forth. In fact, in § IV we show the problem might be so severe that estimating a GMRF-MM using a naive application of GLASSO in the M step can yield such a poor estimate that one might be better off with the baseline GMM (*i.e.*, with no regularization or sparsity), despite the fact that the latter suffers from over-parameterization. To remove the bias effectively we suggest a two-step estimation procedure summarized in Algorithm 2. First, we use GLASSO (without penalizing the diagonal) and use its estimate of the sparsity pattern (but discard the GLASSO estimates of the nonzero values) to define the graphical structure of each component of the mixture (different components can have different sparsity patterns). Then the second step is finding the MLE subject to that support, using the method we described in § III-A. More explicitly, suppose that $\widehat{\boldsymbol{Q}}_k$ is estimated by GLASSO for component $k$. We then remove the $\ell_1$ penalty and estimate $\boldsymbol{Q}_k$ such that $\mathrm{Supp}(\boldsymbol{Q}_k) \subseteq \mathrm{Supp}(\widehat{\boldsymbol{Q}}_k)$. In summary, while this approach requires solving two optimization problems instead of one, it yields substantially better results, as we will show in § IV.

*Remark 6:* Instead of solving a single GLASSO problem, the underlying graphs of the precision matrices can also be estimated by solving $n$ independent (non-Graphical) LASSO problems involving the data terms $\{x_i\}$ [54]. The estimated graphs can then be used as the support for the method described in § III-A. Our experiments indicate that this approach yields comparable results to our approach, both in terms of the estimated graphs and the quality of the model. However, as the data in the EM is softly divided between components, the (rectangular) data matrix can be much larger than the empirical covariance in Eq. (5), and hence obtaining the pattern from the GLASSO estimate is cheaper in that case. Also, using the biased GLASSO estimates, we "warm start" each component's precision-matrix estimation. This results in fewer iterations to convergence.

---

**Algorithm 2** Debiased Graphical LASSO

**Input:** $\boldsymbol{S}, \lambda$
$\widehat{\boldsymbol{Q}} \leftarrow \mathrm{GLASSO}(\boldsymbol{S}, \lambda)$ # *Algorithm initialized by the GLASSO estimate from previous EM iteration.*
$\boldsymbol{Q} \leftarrow$ The result of Algorithm 1 with $\Omega = \mathrm{Supp}(\widehat{\boldsymbol{Q}})$ and $\boldsymbol{Q}^{(0)} = \widehat{\boldsymbol{Q}}$
**return** $\boldsymbol{Q}$

---

#### 1) THE CHOICE OF REGULARIZATION PARAMETER

Just like the standard GLASSO estimation requires a proper choice of $\lambda$, the debiased version requires such a choice too. Obviously, the higher we choose $\lambda$ the sparser the estimated $\boldsymbol{Q}$ will be, *but it will also be more biased*. To truly promote sparsity, $\lambda$ usually has to be substantial, which yields a substantially-biased estimation. Often, as we demonstrate later, the optimal choice of $\lambda$ in GLASSO is essentially a compromise between promoting sparsity and reducing the bias effect in the estimation, leading to less sparsity in the estimation as one may be able to achieve without the bias. In our debiased GLASSO approach, we separate the sparsity promotion and the estimation of the nonzero values. Hence, the optimal value of $\lambda$ for the sparsity estimation is larger than the one for the standard GLASSO alone, because then we are not required to reduce the bias effect.

*Remark 7:* The $\ell_1$ regularization in GLASSO also plays the role of producing some tolerance to noise in the measurements. For this purpose, other simple regularization techniques such as Tikhonov or Riccati regularizations [55] (which do not promote sparsity) are also suitable. However, the amount of regularization needed for having tolerance to noise is usually lower than the regularization needed to promote sparsity; the latter often arises from different considerations, e.g., reducing the number of parameters in the model. Once the sparsity is found, the debiased estimation can also be regularized for the purpose of tolerance to noise in the measurements by $\ell_1$, Tikhonov, or any other suitable regularization approach. In any case, unless the noise level is very high, the noise-related regularization parameter will be much smaller than the one needed for promoting sparsity, leading to a proper noise-related bias in the estimation.

## D. THEORETICAL RESULTS EXPLAINING THE GLASSO BIAS

From a Bayesian perspective, the fact that GLASSO causes a bias is unsurprising: any Bayesian method (the $\ell_1$ penalty can be interpreted in terms of a Laplace-distribution prior) pushes the estimate away from the MLE. Our point, however, is not (the well-known fact) that the bias exists but that it is substantial, arguably more than one might expect (regardless whether the diagonal is penalized or not). In this section we provide new theoretical results that shed some light upon this phenomenon (which, as we mentioned earlier, is even more critical in the mixture setting).

First, we state a simple lemma, showing that if the inverse of the empirical covariance happens to be sparse with the same assumed support as that of the latent $\boldsymbol{Q}_{\text{TRUE}}$ (the true unknown precision matrix), then it is the MLE of $\boldsymbol{Q}_{\text{TRUE}}$.

*Lemma 1:* Suppose that $\boldsymbol{S}^{-1}$ is sparse and that $\Omega = \text{Supp}(\boldsymbol{S}^{-1})$. Then $\boldsymbol{S}^{-1}$ solves Eq. (11).

*Proof:* By the strict convexity of the problem from Eq. (6), in addition to the (convex) linear constraints, we have a unique solution to the sparsity-constrained problem. Since $\text{Supp}(\widehat{\boldsymbol{Q}})$ trivially satisfies the optimality condition and the constraints, then it is the solution. □

The following propositions show that when there is enough data such that $\boldsymbol{S}^{-1}$ is sufficient to estimate the true latent $\boldsymbol{Q}_{\text{TRUE}}$ with high accuracy (so $\widehat{\boldsymbol{Q}}_{\text{MLE}} \approx \boldsymbol{Q}_{\text{TRUE}}$), the GLASSO estimate is still substantially far from $\boldsymbol{Q}_{\text{TRUE}}$, even if the sparsity pattern of the GLASSO estimate is correct.

Let the premise of Lemma 1 hold and let $\boldsymbol{Q}_\lambda$ be the solution of Eq. (12), without penalizing the diagonal. Then, approximately, the off-diagonal entries of $\boldsymbol{Q}_\lambda$ have opposite signs to those of the corresponding entries of $\boldsymbol{S}$. More precisely:

*Proposition 1:* Let $\Omega = \text{Supp}(\boldsymbol{Q}_\lambda)$. Up to a first-order approximation we have that for all $(i,j) \in \Omega$ where $i \neq j$ we have

$$|S_{ij}| > \lambda \Rightarrow \text{sign}(S_{ij}) = -\text{sign}((\boldsymbol{Q}_\lambda)_{ij}). \quad (20)$$

*Proof:* The optimality condition of (12) states that if $(i,j) \in \Omega$

$$(\boldsymbol{Q}_\lambda^{-1})_{ij} = \begin{cases} S_{ii} & i = j \\ S_{ij} + \lambda \cdot \text{sign}((\boldsymbol{Q}_\lambda)_{ij}) & i \neq j \end{cases} \quad (21)$$

and otherwise

$$|(\boldsymbol{Q}_\lambda^{-1})_{ij} - S_{ij}| < \lambda.$$

Let $\boldsymbol{D}_Q$ be the diagonal matrix whose entries are the diagonal of $\boldsymbol{Q}_\lambda$. Then, according to the Neumann series we have

$$(\boldsymbol{D}_Q^{-1}\boldsymbol{Q}_\lambda)^{-1} = \boldsymbol{D}_Q\boldsymbol{Q}_\lambda^{-1} = \boldsymbol{I} - \widetilde{\boldsymbol{Q}}_\lambda + \sum_{k=2}^{\infty}(-1)^k\widetilde{\boldsymbol{Q}}_\lambda^k \quad (22)$$

where $\widetilde{\boldsymbol{Q}}_\lambda$ is the off-diagonal part of the diagonally normalized matrix $\boldsymbol{D}_Q^{-1}\boldsymbol{Q}_\lambda$. Since the diagonal of $\boldsymbol{D}_Q$ is all positive, then using the first-order term of (22) on the right hand side of (21) yields for every $i \neq j$

$$(\boldsymbol{Q}_\lambda^{-1})_{ij} \approx (\boldsymbol{D}_Q^{-1} - \boldsymbol{D}_Q^{-1}\widetilde{\boldsymbol{Q}}_\lambda)_{ij} = S_{ij} + \lambda \cdot \text{sign}((\boldsymbol{Q}_\lambda)_{ij}), \quad (23)$$

which means that any off-diagonal entry $(i,j)$ for which $|S_{i,j}| > \lambda$ is non zero in $\boldsymbol{Q}_\lambda$ $((i,j) \in \text{Supp}(\boldsymbol{Q}_\lambda))$ up to the first-order approximation, and we have

$$(\boldsymbol{D}_Q)_{ii}^{-2}(\boldsymbol{Q}_\lambda)_{ij} = -S_{ij} - \lambda \cdot \text{sign}((\boldsymbol{Q}_\lambda)_{ij}),$$

which proves the proposition for all off-diagonal entries $(i,j)$. □

Suppose that the premise of Preposition 1 holds and assume that we have enough samples such that $\boldsymbol{S}^{-1} \approx \boldsymbol{Q}_{\text{TRUE}}$. Furthermore, assume that with a proper choice of $\lambda$, the support of the GLASSO minimizer $\boldsymbol{Q}_\lambda$ is the true one. This will happen for example when $\boldsymbol{Q}_{\text{TRUE}}$ is diagonally dominant, in which case the entries $S_{ij}$ decay rapidly with the graph distance between $i$ and $j$ on the graph associated with $\boldsymbol{Q}_{\text{TRUE}}$. This can be seen from the corresponding Neumann series in (22). The next proposition shows that even though it has the true support, the eigenvalues of $\boldsymbol{Q}_\lambda$ can be substantially biased; *i.e.*, the GLASSO estimate, even without penalizing the diagonal, can be far from the MLE and the true solution.

*Proposition 2:* Assume that $\boldsymbol{Q}_\lambda$ has the true support; i.e., $\text{Supp}(\boldsymbol{Q}_\lambda) = \text{Supp}(\boldsymbol{Q}_{\text{TRUE}})$. Let $\mu_i$ be the eigenvalues of $\boldsymbol{Q}_\lambda^{-1}$. Then, up to first-order approximation we have that

$$|S_{ii} - \mu_i| \leq \sum_{j \neq i:(i,j)\in\Omega}(|S_{ij}| - \lambda) + \sum_{j \neq i:(i,j)\notin\Omega}(|S_{ij} - \lambda E_{ij}|) \quad (24)$$

where $|E_{ij}| < 1$.

In words: even when GLASSO yields the true support, the eigenvalues of the GLASSO covariance matrix will likely be different from those of the empirical covariance $\boldsymbol{S}$ even if that one approximates the true covariance $\boldsymbol{Q}_{\text{TRUE}}^{-1}$ well. Particularly, the Gershgorin radii are much smaller in comparison to those of $\boldsymbol{S}$ as $\lambda$ grows (while all conditions are satisfied), and the eigenvalues are much closer to the diagonal elements $S_{i,i}$. Moreover, the smallest and largest eigenvalues are expected to be larger and smaller respectively as a result of the bias, indicating that estimated $\boldsymbol{Q}_\lambda$ has an overly-optimistic condition number w.r.t.that of the true matrix. All this while at the same time the proposed sparsity-constrained MLE estimates $\boldsymbol{Q}_{\text{TRUE}}$ well under the stated assumptions, and since the supports are identical, so does the debiased GLASSO estimate.

*Proof:* From the optimality condition in Eq. (21) without penalizing the diagonal, we know that there exists a matrix $\boldsymbol{E}$ such that

$$\boldsymbol{Q}_\lambda^{-1} = \boldsymbol{S} + \lambda\boldsymbol{E}, \quad (25)$$

where $E_{ii} = 0$ for all $i$, and for $i \neq j$ we have that $E_{ij} = \text{sign}((\boldsymbol{Q}_\lambda)_{ij})$ for $(i,j) \in \Omega$, and $|E_{ij}| < 1$ otherwise. From the Gershgorin circle theorem we have that

$$|S_{ii} - \mu_i| \leq \sum_{j \neq i}(|S_{ij} - \lambda E_{ij}|). \quad (26)$$

From Preposition1 we have that up to a first order approximation, if an off-diagonal $(i,j) \in \text{Supp}(\boldsymbol{Q}_\lambda)$, and $|S_{ij}| > \lambda$ then $E_{ij} = -\text{sign}(S_{ij})$. This, together with Eq. (26), concludes the proof. □

We note that the results above will not be significantly different if we consider the original version of GLASSO where the diagonal is penalized.

## IV. RESULTS

In this section, we validate the effectiveness of the proposed model and method via both synthetic- and real-data experiments in several settings. The first two synthetic-data experiments serve to present the GLASSO bias and how it is amplified in the mixture setting, and to demonstrate how the proposed method solves that problem effectively. The first set of real-data experiments are estimations of a single Gaussian for several gene-expression datasets. The second is learning GMMs over image patches, using the different approaches, and then use them as priors within a popular image-denoising method. Whenever we report a certain value of $\lambda$, either for the GLASSO method when used alone, or for the first step in our debiasing method, that value was empirically chosen to yield the best results for the respective method. All the experiments were done using a GLASSO implementation with no penalty on the diagonal elements.

Note that in order to reduce the number of parameters, instead of our GMRF approach, one can also opt to restrict the covariance matrices to be diagonal or to impose a single covariance matrix shared by all the $K$ Gaussians in the model [56]. Those assumptions, however, are more restrictive. For example, our experiments also include cases with zero-mean Gaussians. In such a case the distinction between clusters is done solely by the covariance matrices, so the shared-covariance approach is inapplicable. More generally, in most real-world settings it is fairly unlikely that all of the clusters will have the same shape. Likewise, diagonal covariances cannot capture within-Gaussian correlations across the different dimensions (this is contrast to the GMRF case).

### A. THE GLASSO BIAS

Our first experiment is the task of estimating the precision matrix of a single Gaussian. We drew 300 points, *i.i.d.*, from a zero-mean Gaussian in $\mathbb{R}^{1024}$ whose precision matrix is induced by a homogenous discrete 2D Laplacian operator (for the connection between such a differential operator and a GMRF, see [26]), whose stencil is given by

$$\begin{bmatrix} & -1 & \\ -1 & 4 & -1 \\ & -1 & \end{bmatrix}$$

for a 32-by-32 lattice. The data dimension, $n = 32^2 = 1024$, is larger than the number of samples, $N = 300$, making the empirical covariance matrix rank deficient.

We compare the eigenvalues of the real precision matrix with three of the estimates discussed earlier (in either the GLASSO method or the proposed debiased method, the GLASSO parameter, $\lambda$, is chosen according to the best empirical result for that method); see Figure 1. It is evident from the figure that while both the proposed methods (known sparsity; debiased) obtain eigenvalues that are very close to the correct

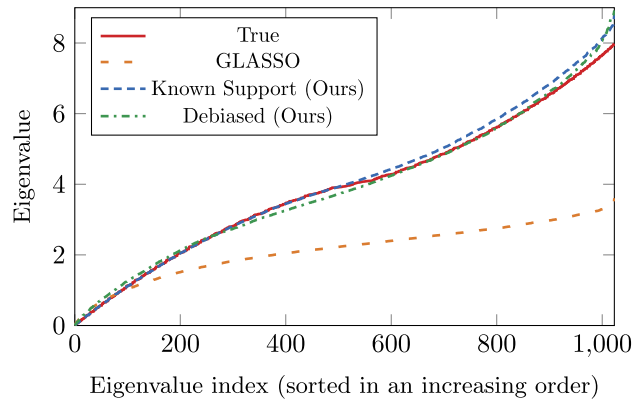

**FIGURE 1.** A comparison of the eigenvalues of the real matrix and the three estimates. $Q_{\text{TRUE}}$ is induced by a 2D discrete Laplacian operator of the dimensions $32^2 \times 32^2$. The estimations were done using 300 samples. Known Support: our MLE estimator given the known support; GLASSO: The GLASSO estimator with $\lambda = 0.25$; Debiased: our MLE estimator with a support chosen to be the nonzero pattern as was estimated by the GLASSO.

ones, the GLASSO tends to yield eigenvalues that are much smaller.

The observed differences show how significant the bias of the GLASSO can be, even in the single-Gaussian case, as well as the substantial improvement in estimation accuracy that is achieved by using the proposed debiasing method.

### B. SYNTHETIC DATA CLUSTERING

Our second experiment focuses on the task of clustering. To generate the synthetic data we draw 10 random sparse SPD matrices and use them as precision matrices of zero-mean Gaussian mixture components. Concretely, each matrix is defined by a finite-difference discretization of an anisotropic diffusion operator

$$\frac{\partial}{\partial x}\left(a(x,y)\frac{\partial}{\partial x}\right) + \frac{\partial}{\partial y}\left(b(x,y)\frac{\partial}{\partial y}\right)$$

on a regular $10 \times 10$ grid, where $a(x, y), b(x, y)$ are the positive diffusion coefficients, chosen at random uniformly. We then draw between 1,500 to 3,000 samples from each component to create a dataset. This way we generated 30 different datasets of random components.

For each dataset we use the EM-GMM algorithm described in § II to learn GMMs of several types, each is different only in the precision-matrix estimation procedure: 1) a baseline with no regularization (*i.e.*, $Q = S^{-1}$); 2) GLASSO estimator with $\lambda = 0.3$; 3) The proposed debiasing method with the same parameter; 4) The proposed method using the known support. Note that the value of $\lambda$ was chosen according to GLASSO's best empirical result. Table 1 shows the mean and standard deviation of the Normalized Mutual Information (NMI) and Variation of Information (VI) achieved by the different configurations on the datasets.

To summarize, the best score is of the proposed GMRF-MM using the known sparsity, showing this is the best choice when the graphical structure underlying the data is known. The GLASSO achieves a very poor result; in fact, it is even (far) worse than the baseline GMM, even though the latter is over-parameterized. In contrast, the proposed debiased method greatly outperforms the baseline, and achieves results that are almost equal to those obtained when we use the known sparsity.

**TABLE 1.** Synthetic-data clustering.

| GMM type | NMI (higher is better) | VI (lower is better) |
|---|---|---|
| Baseline | $0.61 \pm 0.15$ | $1.78 \pm 0.67$ |
| GLASSO | $0.03 \pm 0.02$ | $2.77 \pm 0.18$ |
| Debiased (Ours) | $0.92 \pm 0.01$ | $0.39 \pm 0.12$ |
| Known Sparsity (Ours) | $\mathbf{0.94 \pm 0.02}$ | $\mathbf{0.29 \pm 0.09}$ |

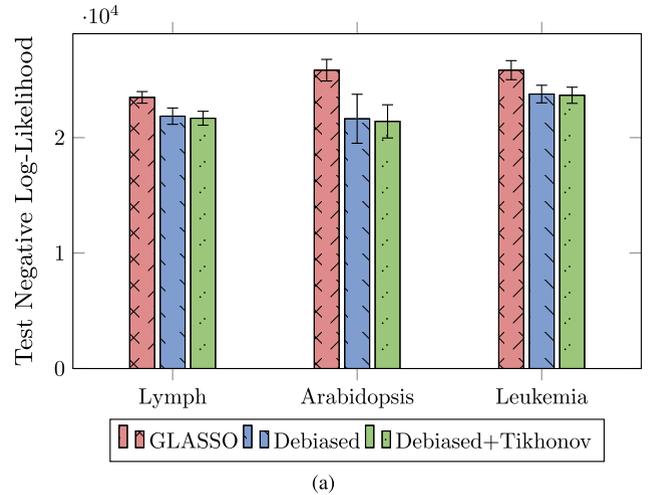

### C. GENE EXPRESSION DATA

The first real-world data we experiment with are gene-expression datasets reported in [55], [57]. In this application domain, estimating precision matrices is of interest as it facilitates the discovery of interactions between variables (gene expressions) in high-dimensional datasets. These datasets have many variables and very few samples, making the empirical covariance rank deficient. In this experiment, we compare GLASSO and our debiased method in estimating a single Gaussian for modeling each dataset.

Each dataset is preprocessed so each variable is of zero mean and unit variance. Then, we perform 30 repetitions, where in each repetition we randomly split the data to a train set (80%), and a test set (20%). Similarly to [30], [55], [57], the regularization parameter $\lambda$ for GLASSO is chosen for each dataset to keep the number of nonzero entries in the precision matrix to be about 10 nonzeros per row and we compare the likelihood of the obtained model under this constraint. We use GLASSO, Debiased GLASSO, and Debiased GLASSO with a Tikhonov regularization with a small parameter. The latter is important since these are real data with a significant amount of noise (see Remark 7).

The first three datasets – Lymph, Arabidopsis, and Leukemia – have sample dimensions 587, 834, and 1255 respectively, and the number of samples in each is 147, 117, and 71. The results are presented in 2a. Because of the debiasing effect we can see 10%–20% improvement in the negative log-likelihood over that obtained by the GLASSO.

The next datasets we use are taken from the Gene Expression Omnibus.[1] The dimensions of each sample are 20K-100K, and there are about 75–280 samples in each set. Similarly to [55], in each repetition, we randomly choose 700 features. The results are presented in 2a, and again show an advantage to the debiased estimates. We note that for this experiment since we choose only 700 features from each sample, we can choose quite low values of $\lambda$ (and $\lambda \approx 0.5$). Having more features generally requires higher values of $\lambda$ to yield sparsity of 10 nonzeros per row. Specifically, the full data requires values $\lambda \approx 0.7$ [30], further increasing the bias.

[1] https://www.ncbi.nlm.nih.gov/geo/

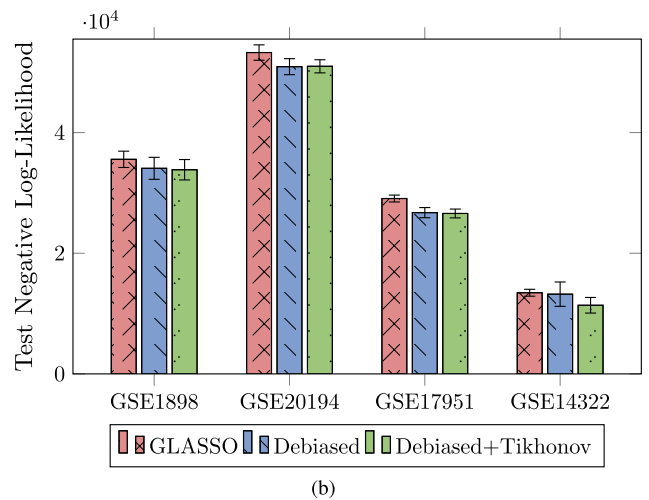

**FIGURE 2.** Gene Expression mean negative log-likelihood (with standard deviation interval) for the test data. Lower is better. Comparing GLASSO with our Debiased and Debiased with Tikhonov regularization methods. (a) uses all the features of each dataset, while (b) uses 700 features chosen randomly for each test.

### D. IMAGE RESTORATION WITH GMM PRIOR

In the following experiments, we test the quality of our method for learning a prior for the task of image denoising. Particularly, we use an image-restoration method from [6] which is based on maximizing the Expected Patch Log-Likelihood (EPLL) while being close to the corrupted image. The restoration is achieved by minimizing the following objective function

$$f_p(\boldsymbol{x}|\boldsymbol{y}) = \frac{\alpha}{2}\|\boldsymbol{x} - \boldsymbol{y}\|^2 - \text{EPLL}_p(\boldsymbol{x}) \quad (27)$$

w.r.t. $\boldsymbol{x}$, where $\boldsymbol{y}$ is the noisy image and $\alpha > 0$ is a balancing parameter. The EPLL is defined as

$$\text{EPLL}_p(\boldsymbol{x}) = \sum_i \log p(\boldsymbol{P}_i \boldsymbol{x}) \quad (28)$$

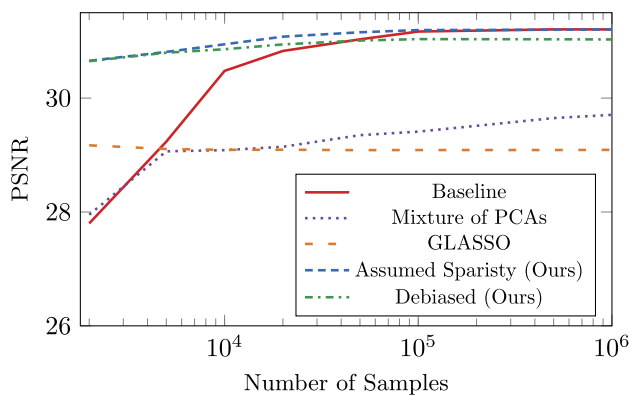

**FIGURE 3.** Performance of the EPLL-based image restoration method with each of the estimated GMMs as prior, using varying numbers of training samples. The abscissa represents the number of samples used to train each model, and the ordinate represents the mean PSNR result (higher is better) for denoising 200 test images corrupted with a white Gaussian noise of $\sigma = 15$.

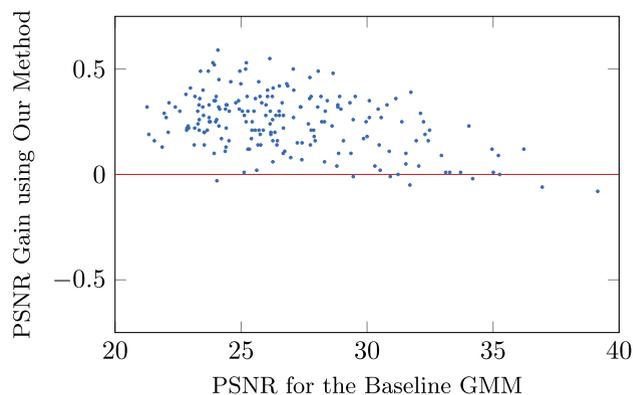

**FIGURE 4.** Comparison of the performance of RGB image denoising using the EPLL [6], extended from grayscale to color images. The two priors are a baseline GMM, and a GMRF-MM with an assumed sparsity. Both priors are learned using $10^6$ patches of size $8 \times 8 \times 3$, with $K = 200$. Each image was corrupted with a white Gaussian noise of random $\sigma$.

where $P_i$ is a projection matrix that extracts the $i$-th patch from the image. The prior $p$ proposed by [6] is a GMM learned over the pixels of natural image patches. In this experiment we use it as a "black box" and change only the GMM priors. We use the Peak Signal-to-Noise Ratio (PSNR) between the clean images and the reconstructions as a means to measure the quality of each GMM type in modeling natural image patches.

For this, we have implemented a version of the EPLL method in Python, which was adapted from the Matlab code of [6], which was written for grayscale images, to also work with RGB images. We set the restoration method to run 5 iterations for each image. Each experiment specifies the GMMs learning configurations. All of the learning tasks were done using randomly-selected patches taken from the Berkeley Segmentation Dataset [58] (BSD) training set, and all tests were done on the BSD test set.

#### 1) GRAYSCALE IMAGES

The first experiment in this set is for grayscale images. Treating the $d \times d$ patch as a graph with $n = d^2$ vertices, we define a neighborhood system that connects all vertices in an $m \times m$ region. As explained in § II, this creates a sparsity pattern for each component's precision matrix. Enforcing such sparsity reduces the number of nonzero parameters from $O(d^4)$ to $O(d^2 m^2)$.

For this experiment we set $K$, the number of Gaussians, to 100, and change only the size of the training set each time. The parameter $\lambda$ was picked according to the best empirical result for the smallest training set, and kept while increasing the set size. We use our EM-GMM implementation to learn several GMMs, each is different only in the precision matrix estimation process: 1) A baseline GMM with no regularization (i.e., $Q = S^{-1}$); 2) A mixture of 30-dimensional Principal Component Analysis (PCA) subspaces [59]; 3) The GLASSO estimator with $\lambda = 0.7$; 4) The proposed debiased method with the same parameter; 5) The proposed method using the sparsity pattern described above with $m = 5$.

To test the quality of each of the GMMs we use the denoising algorithm described above to denoise the 200 images of the BSD test set corrupted with a white Gaussian noise of $\sigma = 15$. Figure 3 shows the mean PSNR result of each GMM type as a function of the size of the training set. The figures show that the smaller the training set is, the worse the results of the baseline GMM are, and that the proposed method using the assumed sparsity clearly outperforms it. As the number of samples grows, the baseline GMM and our GMRF-MM achieve comparable performance, *even though our approach has far fewer parameters*. Also, we see that the proposed debiasing method achieves better results than GLASSO. This shows that when we cannot assume the graphical structure of the data, our debiasing method is still preferable over GLASSO. Results for other noise levels (values of $\sigma$) showed a similar trend.

#### 2) COLOR IMAGES

The next experiment we present is for RGB images, where each pixel has 3 color channels. In this scenario, each patch is of size $d \times d \times 3$. We define a neighborhood system such that each pixel is connected to the $m \times m$ pixels over all of the color channels. We set $K$ to 200, and learn the four models using $10^6$ RGB patches taken randomly from the BSD training set. We learn two types of GMMs, one with an assumed sparsity using $m = 5$, and a baseline GMM with no regularization. Each image from the BSD test set is first corrupted using a random $\sigma$ drawn uniformly in the range of 10 to 100 and is then denoised with each of the learned priors. Earlier, with grayscale images, we saw that $10^5$ samples were about enough for the baseline GMM to achieve similar results to the proposed GMRF-MM. Here, when the dimensionality is higher we show that the proposed model is more effective than the baseline even when their common training set is larger: Figure 4 shows the performance gain from using the proposed method. Every point above the line represents an image, corrupted with random noise, on which a better PSNR result was obtained when denoised with our GMRF-MM. The gain is fairly consistent, even though our model uses far fewer parameters.

## V. CONCLUSION

In this work, we proposed a GMRF mixture model to mitigate the over-parameterization effect that classical GMMs are prone to in the presence of data scarcity. We presented an effective way to learn such a model in two cases: one is when the sparsity pattern of the precision matrices is assumed, and the other is when that pattern needs to be estimated. The latter is the more complicated scenario, for which we showed that a naive approach of incorporating the GLASSO estimator in the traditional GMM framework leads to unfavorable models due to a biasing effect of the $\ell_1$ regularizer in GLASSO. To overcome this, we proposed a two-stage estimation procedure where in the first stage only the sparsity pattern is estimated via GLASSO, while in the second stage the sparsity is held fixed, the $\ell_1$ regularization is reduced, and a Newton-type optimization method is used to find the globally-optimal precision matrix. We showed in our numerical results that in many scenarios the baseline standard GMM can produce unfavorable models, and restricting its parameters can improve the estimated models. We also showed that the model produced by GLASSO is consistently improved by the debiasing stage, and in several cases can outperform the standard GMM - especially when the data is insufficient for effective learning of the latter.

*We reiterate that our main motivation is related to the mixture-model case*, where both the over-parameterization and the LASSO bias are far more drastic (than they are with a single Gaussian); this case was not well studied before. Thus, the success of the novel proposed method in the single-Gaussian case should not detract the reader from appreciating its more important success in the mixture-model setting.

Our framework can be especially effective in scenarios where the dimensions are high, the underlying distribution has a large number of modes, and the data is scarce w.r.t. the number of parameters associated with the multimodal distribution. Another interesting utility of our method, which we did not explore here, is to exploit the sparsity and compactness of the learned GMRF-MM to gain computational benefits during inference tasks where that model is used. For example, the EPLL framework for image restoration might be able to leverage computational tools from sparse linear algebra to speed up computations and reduce its memory footprint.

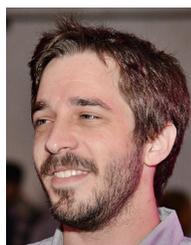

**SHAHAF E. FINDER** received the B.Sc. and M.Sc. degrees in computer science from the Ben-Gurion University of the Negev, Be'er Sheva, in 2018 and 2020, respectively, where he is currently pursuing the Ph.D. degree with the Department of Computer Science.

His research interests include computer vision, machine learning, and optimization.

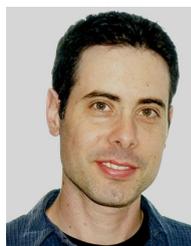

**ERAN TREISTER** received the Ph.D. degree in computer science from the Technion—Israel Institute of Technology, Haifa, Israel, in 2014. He was a Post-Doctoral Researcher in computational geophysics with The University of British Columbia (UBC), Vancouver, BC, Canada. He is currently an Assistant Professor with the Faculty of Computer Science, Ben-Gurion University of the Negev (BGU), Beer-Sheva, Israel. His research interests include multi-scale computational techniques, scientific computing, computational physics and geophysics, image processing and analysis, optimization methods, machine learning, and deep learning.

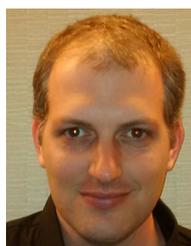

**OREN FREIFELD** received the B.Sc. and M.Sc. degrees in biomedical engineering from Tel Aviv University, in 2005 and 2007, respectively, and the M.Sc. and Ph.D. degrees in applied mathematics from Brown University, in 2009 and 2013, respectively. He was a Postdoctoral Researcher with the MIT Computer Science and Artificial Intelligence Laboratory. He is currently an Assistant Professor with the Department of Computer Science, Ben-Gurion University of the Negev. His main research interests include computer vision, statistical inference, and machine learning.

○ ○ ○